\documentclass[10pt,journal,compsoc]{IEEEtran}

\ifCLASSOPTIONcompsoc
  \usepackage[nocompress]{cite}
\else
  \usepackage{cite}
\fi

\usepackage{lineno,hyperref}
\usepackage{amsmath,graphicx}
\usepackage{amssymb,mathrsfs}
\usepackage{epstopdf}
\usepackage{bbold}
\usepackage{hyperref}
\usepackage{epstopdf}
\usepackage{graphicx}
\usepackage{graphics}
\usepackage{epstopdf}
\usepackage{multirow}
\usepackage{color,soul}
\usepackage{bm}
\usepackage{enumitem}

\ifCLASSINFOpdf

\else

\fi

\begin{document}

\title{ StressedNets: Efficient Feature Representations via Stress-induced Evolutionary Synthesis of Deep Neural Networks }

	\author{	Mohammad Javad Shafiee, \textit{IEEE Member}, Brendan Chwyl, Francis Li, Rongyan Chen,  Michelle  Karg, \mbox{Christian Scharfenberger, and Alexander Wong, \textit{IEEE Senior Member}}
	\IEEEcompsocitemizethanks{\IEEEcompsocthanksitem M.~J. Shafiee, B. Chwyl, F. Li, and A. Wong are with the Department of
Systems Design Engineering, University of Waterloo, Waterloo, Canada.\protect
\IEEEcompsocthanksitem R. Chen, M. Karg, and C.  Scharfenberger are with ADC Automotive Distance Control Systems GmbH, Continental, Germany.}}

\markboth{}%
{Shafiee \MakeLowercase{\textit{et al.}}: StressedNets}

\IEEEtitleabstractindextext{%
\begin{abstract}
The computational complexity of leveraging deep neural networks for extracting deep feature representations is a significant barrier to its widespread adoption, particularly for use in embedded devices.  One particularly promising strategy to addressing the complexity issue is the notion of evolutionary synthesis of deep neural networks, which was demonstrated to successfully produce highly efficient deep neural networks while retaining modeling performance.  Here, we further extend upon the evolutionary synthesis strategy for achieving efficient feature extraction via the introduction of a stress-induced evolutionary synthesis framework, where stress signals are imposed upon the synapses of a deep neural network during training to induce stress and steer the synthesis process towards the production of more efficient deep neural networks over successive generations and  improved model fidelity at a greater efficiency. The proposed stress-induced evolutionary synthesis approach is evaluated on a variety of different deep neural network architectures (LeNet5, AlexNet, and YOLOv2) on different tasks (object classification and object detection) to synthesize efficient \textbf{StressedNets} over multiple generations.  Experimental results demonstrate the efficacy of the proposed framework to synthesize StressedNets with significant improvement in network architecture efficiency (e.g., 40$\times$ for AlexNet and 33$\times$ for YOLOv2) and speed improvements (e.g., 5.5$\times$ inference speed-up for YOLOv2 on an Nvidia Tegra X1 mobile processor).
\end{abstract}

\begin{IEEEkeywords}
deep neural networks, stress-induced evolutionary synthesis, evolutionary synthesis.
\end{IEEEkeywords}}

\maketitle

\IEEEdisplaynontitleabstractindextext
\IEEEpeerreviewmaketitle

\ifCLASSOPTIONcompsoc
\IEEEraisesectionheading{\section{Introduction}\label{sec:introduction}}
\else
\section{Introduction}
\label{sec:introduction}
\fi

\IEEEPARstart{F}{eature} learning, representation, and extraction is a crucial and challenging aspect of machine learning and computer vision as the modeling performance is heavily dependent on the extracted features.  While a very large number of approaches have been proposed to tackle this challenge, it is still considered as an open problem.  At a high level, feature representation approaches can be primarily divided into two main categories: i) hand-crafted feature representations, and ii) learned feature representations.

Hand-crafted feature representations, which are quantitative features that are designed by human experts, have a long history in machine learning and computer vision literature. For instance, histogram of oriented gradient (HOG) features had been utilized in different problems such as people detection~\cite{dalal2005histograms}, crowd segmentation~\cite{tu2008unified} and object detection~\cite{dalal2006object}. Scale-invariant feature transform (SIFT)~\cite{lowe2004distinctive} or speeded up robust features (SURF)~\cite{bay2006surf} have been applied in a wide variety of  applications such as image matching~\cite{grauman2005efficient}, object classification~\cite{schmitt2011object}, image registration~\cite{ayers2007home} and action recognition~\cite{scovanner20073}. In addition to these well-known feature representations,  there have been a wide variety of other hand-crafted feature representations for different applications from different edge detection algorithms~\cite{canny1987computational}, corner detection~\cite{rosten2006machine}, texture extraction~\cite{smith1996local}, just to name a few.  While the hand-crafted features designed by human experts has shown good performance in different problems, they are often limited to being applicable specific applications and highly dependent on the knowledge of human experts.

To mitigate the high dependency on the knowledge of human experts to hand craft features for particular applications, learned feature representations are automatically learned directly from training data using machine learning approaches~\cite{ye2012unsupervised,boureau2008sparse}. Such learned feature representations are typically learned simultaneously with the training of the inference approach designed for a particular task (e.g., for a classification task, the feature extractor and classifier are learned and trained at the same time), and thus the learned features can be considered to be optimal in an error minimization sense for the inference task at hand.

One of the most successful strategies in recent years for feature learning is deep learning~\cite{ramos2012texture}, where the feature extraction and inference method for a particular task can be learned and performed within an end-to-end learning process.  In particular, deep neural networks~\cite{hannun2014deep,krizhevsky2012imagenet,lecun2015deep}, the most popular form of deep learning, have demonstrated tremendous success for learning powerful feature representations from data, leading to state-of-the-art performance in a variety of different applications over the past decade such as object detection~\cite{YOLO2017,ren2015faster}, semantic image segmentation~\cite{badrinarayanan2015segnet,long2015fully}, image classification~\cite{krizhevsky2012imagenet,simonyan2014very}, speed recognition~\cite{hannun2014deep,hinton2012deep}, and gene sequencing~\cite{alipanahi2015predicting}.

The success of deep neural networks for learning feature representations is influenced by two important factors. First of all, a deep neural network training process can be formulated as an end-to-end approach where feature extraction and inference are trained simultaneously based on training data. This way makes the learning in different layers of deep neural networks as a joint process where the best possible feature representation is optimized through the training step. This approach helps deep neural network to be optimized in the inference time as well. For example,  the convolution layers and the fully connected layers of convolutional neural networks~\cite{krizhevsky2012imagenet}, are designed to play the roles of  feature extractor and the inference framework respectively. This type of configuration is very effective since each part (i.e., feature extractor or inference framework (classification))  can compensate for the modeling deficiencies of each other when trained together and leads to an efficient feature representation. Given the rise in big data, deep neural networks can learn highly powerful feature representations around such data for very high inference performance.

Secondly, the significant growth of computational power, particularly the rise in parallel computing devices such as graphics processing units (GPUs) and distributed computing systems, has greatly accelerated the training and inference speed of deep neural networks.  For example, the seminal paper of \mbox{Krizhevsky {\it et al.}~\cite{krizhevsky2012imagenet}} described a new approach for enabling the training of deep neural networks on a GPU which is the turning point of deep neural networks. These improvements in high-performance computing devices have encouraged  researchers to focus on the design of larger and deeper neural networks~\cite{simonyan2014very} that can produce more and more powerful feature representations.

Despite the successes demonstrated with the design of more complex deep neural networks, one of the main drawbacks to this approach is that the improvement in performance often came at the expense of increased complexity, making such networks not well-suited for many applications, particular those that require efficient inference on embedded systems with considerable computational and memory limitations such as self-driving cars, smart-phone applications, and surveillance cameras.  For example, high-performance yet complex deep neural networks are not only more computationally expensive but also require large memories to store an enormous number of network parameters. Fast data transmission is required additionally to support the expensive computation and to load the large network parametrization. These issues associated with computational complexity, memory complexity and bandwidth can be considered as the main barriers to widespread adoption of deep neural networks for feature extraction in a variety of operational scenarios and applications.

To tackle the challenge of computational complexity when leveraging deep neural networks for learning and extracting feature representations, there has been a very strong recent interest towards obtaining efficient deep neural networks capable of producing efficient deep features~\cite{lecun1989optimal,gong2014compressing,han2015deep,guo2016dynamic,jaderberg2014speeding,ioannou2015training,SSL_2016,molchanov2017variational,molchanov2017variational}.
One particularly promising strategy to addressing the complexity issue is the notion of evolutionary synthesis of deep neural networks (EvoNet)~\cite{EvoNet1,shafiee2016evolutionary,shafiee2017learning}, where biological processes are mimicked within a probabilistic framework to synthesize progressively more efficient deep neural networks generation after generation.  This evolutionary synthesis strategy was demonstrated to successfully produce highly efficient deep neural networks while retaining modeling performance, thus enabling efficient yet powerful deep feature extraction.

In the most recent work by Shafiee {\it et al.}~\cite{shafiee2017learning}, they took inspiration from a study by \mbox{Dias \&  Ressler}~\cite{dias2014parental}, which studied the inheritance of parental traumatic exposure~\cite{shafiee2017learning} to their offsprings and found that environmental stimuli imposed on the exposed parents -- here,  an olfactory traumatic exposure on mice-- had a strong genetic influence on their offsprings that were not conceived at the time. This fascinating effect has been also showed by Klosin {\it et al.}~\cite{klosin2017} where environmental information, induced by environmental stresses experienced during the lifetime of C. elegans, was transmitted genetically to subsequent generations.  Inspired by how past stressful experiences are passed down through genetics from generation to generation,
Shafiee {\it et al.}~\cite{shafiee2017learning} mimicked this phenomena in their preliminary work by imposing environmental stresses on an ancestor network during training, so that it results in a genetic encoding favoring the synthesis of even more efficient and robust offspring networks.  The preliminary results presented in~\cite{shafiee2017learning}, where evolutionary synthesis was performed on the AlexNet architecture~\cite{krizhevsky2012imagenet} for the task of image classification, showed considerable promise in synthesizing highly efficient deep neural networks with strong model accuracy retainment, an thus motivates us to extend upon that preliminary work to fully develop an evolutionary synthesis framework built around environmental stress induction.

Motivated by the encouraging results, we further extend upon the initial ideas presented in~\cite{shafiee2017learning} via the introduction of a formalized stress-induced evolutionary synthesis framework, where stress signals are imposed upon the synapses of a deep neural network during training to induce stress and steer the synthesis process towards the production of more efficient deep neural networks (which we will refer to as \textit{StressedNets} over successive generations.  This stress-induced evolutionary synthesis approach improves the robustness of the synthesized network architectures in facing traumatic changes, which as a consequence promotes the synthesis of StressedNets with improved model fidelity at a greater efficiency.  More specifically, in the formalized stress-induced evolutionary synthesis framework, the training of deep neural networks within an evolutionary synthesis framework is formulated as a maximum a posteriori (MAP) problem, with traumatic stresses to synapses encoded within the prior model.  The prior model is designed such that the distribution of synaptic strength in an exposed parent deep neural network is tailored to exhibit inherent genetic encodings to favor offspring neural networks with greater efficiency during the synthesis process, thus transmitting the environmental information experienced by a deep neural network from generation to generation.

While the initial idea was introduced in~\cite{shafiee2017learning}, this paper makes the following significant contributions to greatly expand beyond the initial introduction of the proposed framework  in~\cite{shafiee2017learning}:
\begin{itemize}
	\item More comprehensive formalization and discussion of the stress-induced evolutionary synthesis process within the MAP framework,
	\item introduction and evaluation of a large family of StressedNets based on a variety of different network architectures (LeNet5, AlexNet, and YOLOv2) on different types of tasks (object classification and object detection) to demonstrate generalizability of the proposed approach,
\item more comprehensive evaluation of stress-induced evolutionary synthesis using a wider variety of benchmark datasets (i.e., MNIST, CIFAR-10, KITTI) to demonstrate generalizability of the proposed approach,
\item introduction of performance comparison with six different state-of-the-art methods for achieving efficient deep neural networks (i.e.,~\cite{han2015deep,guo2016dynamic,ullrich2017soft,molchanov2017variational,louizos2017bayesian}),
\item introduction of a run-time evaluation of stress-induced evolutionary synthesis on an embedded processor (i.e., Nvidia Tegra X1),
	\item introduction of a comprehensive parametric test to study the effect of the environmental factor parameter in evolutionary deep intelligence framework, and
	\item introduction of a comprehensive qualitative and quantitative feature analysis on the generated features within StressedNets.
\end{itemize}

The paper is organized as follow. In Section~\ref{sec:related}, related work in achieving efficient deep neural networks is presented to provide context.  In Section~\ref{sec:method}, the proposed stressed-induced evolutionary synthesis framework is formalized and explained in a detailed manner. In Section~\ref{sec:exp}, the proposed stressed-induced evolutionary synthesis framework is comprehensively examined and evaluated on a variety of different deep neural network architectures (LeNet5, AlexNet, and YOLOv2), on different tasks (object classification and object detection), and different benchmark datasets (MNIST, CIFAR-10, KITTI) to synthesize efficient \textbf{StressedNets} over multiple generations, along with a study of the effect of environmental factors as well as quantitative and qualitative feature analysis.

\section{Related Work}
\label{sec:related}

Prior to discussing the proposed stressed-induced evolutionary synthesis framework in great detail, it is important to give context to previous related methods for achieving efficient deep neural networks from which deep feature representations can be obtained.  The majority of methods in previous literature on achieving efficient deep neural networks can be grouped into two main categories:  I) methods addressing memory complexity associated with deep neural networks, and II) methods focusing on computational and memory complexity issues together.

In the area of methods tackling memory complexity, Lecun {\it et al.}~\cite{lecun1989optimal} addressed this issue in their seminal paper by proposing the optimal brain damage method where synapses were pruned based on their strengths. They utilized the second-derivative information to specify the neuron to be pruned and made a trade-off between the number of parameters and training error. They formulated a new error function and the effect of perturbing the parameter vector was analytically calculated during the training. The main framework can be summarized as follow, the initial network architecture is chosen and the network is trained to obtain a reasonable performance. Based on the second derivative the saliency value for each parameter is computed and at then end, the set of parameters with low-saliency value are removed from the network model. The proposed approach took advantage of information theory to select non-important parameters in the model to be removed. The neural networks can be considered as a non-linear mapping between inputs and outputs where the network parameters extract the knowledge, therefore, different information theoretic methods can be applied in this area.

 Gong {\it et al.}~\cite{gong2014compressing} took advantage of information-theoretical vector-quantization methods to compress the parameters of the network. They used k-means clustering on the weights to quantize the parameters of the dense connected layers. They examined different quantization algorithms in different levels including binarization, vector quantization,  product quantization and residual quantization and compared them against each in terms of saving storage requirement of deep neural network given the preservation of the modeling accuracy to some extent.

  To further reduce the network structure and the storage requirement, Han {\it et al.}~\cite{han2015deep} proposed the combination of pruning, quantization and Huffman coding.
Followed by optimal brain damage approach~\cite{lecun1989optimal}, the weights with smaller weights and below a pre-defined threshold are pruned from the network and the network is trained again to compensate for the loss. Then they applied a quantization and weight sharing approach to reduce the number of required bits to store a weight in the network. They also performed Huffman coding to further reduce the bit storage based on the occurrence of each weights in the network. Guo {\it et al.}~\cite{guo2016dynamic} extended upon this algorithm and proposed dynamic network surgery  method where beside pruning the splicing procedure is performed. The splicing procedure enables connection recovery once the pruned connections are found to be important.

 The storage demand of deep neural networks is one issue needed to be resolved, however the bigger issue is the computational complexity and running time problem when deep neural networks are processed on embedded devices where several methods have been trying to address this issue.
In the area of methods addressing computational and memory complexity issues simultaneously, low-rank matrix factorization~\cite{jaderberg2014speeding,ioannou2015training} was proposed to approximate the filter structures and convolutional kernels in convolutional layers.  For example, Jaderberg {\it et al.}~\cite{jaderberg2014speeding} took advantage of low-rank matrix factorization to learn separable smaller kernels like~\cite{ioannou2015training}, the separable kernels are optimized after training the network. The convolutional kernels are approximated based on a filter banks of  horizontal and vertical kernels. The proposed filter bank approaches reduces the redundancy among filters by approximate them vai smaller kernel playing the role of bases.   By use of this approach, in addition to reduce the number of parameters, they could decrease the computational complexity and as a results, decreased the running time of the feed-forward pass through the network.

  Ioannou~{\it et al.}~\cite{ioannou2015training} proposed a new training approach such that the network learns a set of small basis filters from scratch via low-rank matrix factorization and by
  using smaller kernel size addresses the running-time issue. The learned kernels are rectangular in the spatial domain. The conventional squared kernels ($k\times k$) are factorized into rectangular horizontal ($1\times k$) and vertical kernels ($k\times1$) which their responses are then linearly combined by the next layer of $1\times1$ filters.

Learning the structures of kernels during the training process is another way to address the network optimization issue.
 Wen~{\it et~al.}~\cite{SSL_2016} suggested applying regularization techniques to learn the kernel structures and account for structured sparsity. They introduced a new regularization approach to learning the filter shapes and layer depth during training. They formulated the loss function to account for the structure of the network as well. The proposed loss function is the combination of loss on data, a non-structured regularization on every weight in the network and a structured sparsity regularization on each layer. They applied a group Lasso on a set of weights which can zero out all weights in the set. The proposed approached tends to remove less important filters and channels in the network.

 Variational learning and Bayesian algorithms~\cite{ullrich2017soft,molchanov2017variational} are other techniques that have been proposed to formulate the model compression and network optimization. Ullrich {\it et al.}~\cite{ullrich2017soft} took advantage of minimum description length in a variational learning framework for neural network compression. They enforced  the sparsity and model compression via a prior distribution during the training time. Molchanov {\it et al.}~\cite{molchanov2017variational} utilized a variational dropout approach to sparsify the neural networks. They applied  an unbounded dropout technique leading to sparse neural networks.

 Louizos {\it et al.}~\cite{louizos2017bayesian} extended upon the proposed method by Ullrich {\it et al.}~\cite{ullrich2017soft} and used a hierarchical priors to prune neurons instead of synapses in the network. They applied a sparsity inducing priors for hidden units instead of  individual weights which prunes neuron instead of synapses in the network. They also utilized a posterior uncertainty to determines the optimal fixed point precision.

\begin{figure*}[!ht]
	\begin{center}
		\includegraphics[width = 18 cm]{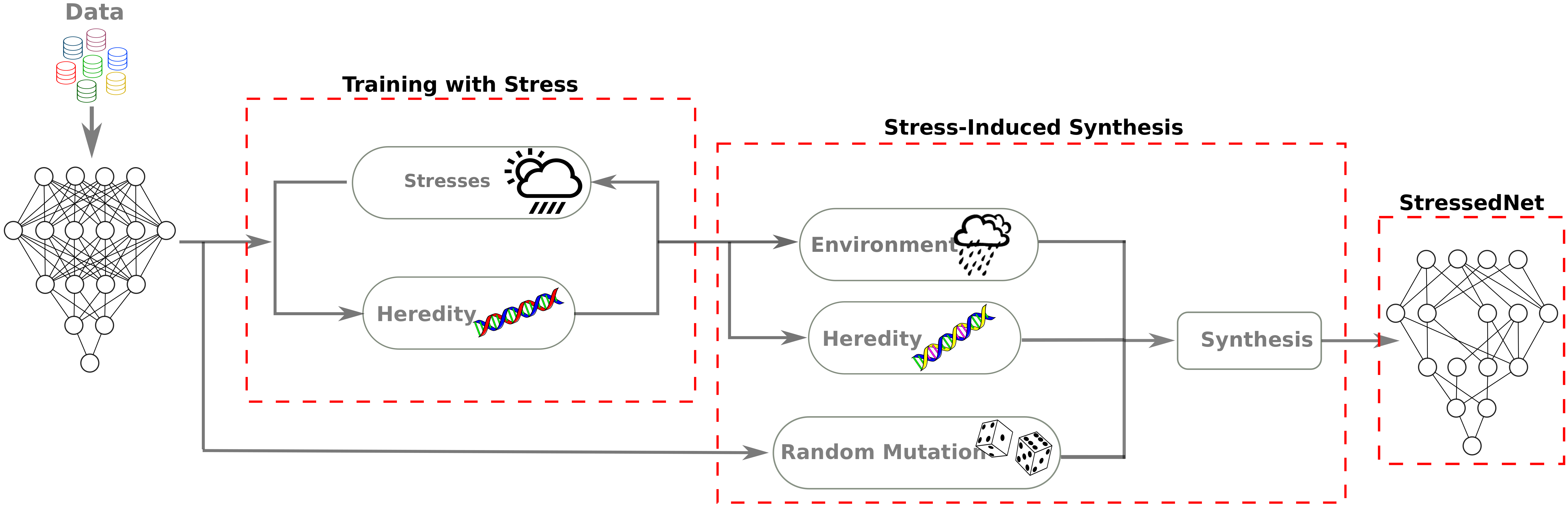}
	\end{center}
	\caption{Overview of the proposed stress-induced evolutionary synthesis framework (for simplification purposes, only a single generation is shown). The architectural traits of ancestor networks are encoded via probabilistic `DNA' sequences. Environmental stresses are induced during each epoch of training, thus enabling synapses to be prepared for a traumatic exposure between generation.  A new offspring \textit{StressedNet} is synthesized in each generation based on the probabilistic `DNA' sequences (heredity), environmental factors (natural selection), and random mutation. The main steps of the proposed framework are I) induce stress during the training process (i.e., training with stress), II) synthesize StressedNets based on the probabilistic DNA of the ancestor network trained with stress (i.e., stress-induced synthesis). The proposed framework synthesizes a new StressedNets with network architectures that are more efficient compared to its ancestor.}
	\label{fig:flowdiagram}
\end{figure*}

Another promising approach to tackling both the computational and memory complexity issues simultaneously is the evolutionary sythesis (EvoNet) framework proposed by Shafiee~{\it et~al.}~\cite{EvoNet1}, where inspirations from evolutionary biology such as random mutation, natural selection, and heredity were leveraged within a probabilistic framework to synthesize increasingly efficient deep neural networks over successive generations, resulting in the learning of highly efficient yet powerful feature representations. While previous works have explored the use of evolutionary computing methods for training and generating deep neural networks~\cite{stanley2002evolving,white1993gannet}, they have not only largely focused on accuracy and not on progressively more efficient deep neural networks, but also have leveraged classical methods such as genetic algorithms and evolutionary programming which differs greatly from the probabilistic generative framework proposed in~\cite{EvoNet1}.

One of the key aspects of the evolutionary synthesis framework greatly influencing the efficiency and quality of the synthesized offspring deep neural networks is the genetic encoding scheme, which acts as a probabilistic `DNA' to mimic the heredity aspect of biological evolution. For instance, \mbox{Shafiee \& Wong}~\cite{shafiee2016evolutionary} extended the genetic encoding scheme to synthesize deep neural networks with architectures that enable more efficient inference on parallel computing devices such as GPUs.  More specifically, they proposed a new genetic encoding scheme to promote the formation of highly sparse sets of synaptic clusters, thus tailoring them to the hardware architecture of GPUs that can execute a set of kernel computing instructions in a highly parallel manner.

\section{Methodology}
\label{sec:method}

Here, we introduce and formalize an extended evolutionary synthesis framework for learning efficient deep feature representations by using stress-induced evolutionary synthesis strategy where several synapses in a network are exposed to stress signals during the training to induce stress within the network. The imposed stress signals leveraged here for inducing environmental stresses improves the robustness of the synthesized network architectures in facing traumatic changes, which as a consequence promotes the synthesis of StressedNets with improved model fidelity at a greater efficiency.  In this section, we first review the underlying concept behind evolutionary synthesis of deep neural networks, followed by a detailed description and explanation of the proposed stress-induced evolutionary synthesis scheme.

\subsection{Evolutionary Synthesis of Deep Neural Networks}

The evolutionary synthesis framework leveraging in this work was first proposed by Shafiee {\it et al.}~\cite{EvoNet1}, where progressively more efficient deep neural networks are synthesized within a probabilistic framework over multiple generations by leveraging processes that mimic heredity, natural selection and random mutation.  More specifically, the architectural traits of a deep neural network are modeled by synaptic probability models that can be considered as the probabilistic `DNA', and that are used to mimic heredity to pass genetic information to subsequent generations. Offspring deep neural networks with diverse network architectures are synthesized stochastically based on this probabilistic `DNA' together with probabilistic computational environmental factor models for encouraging progressively increasing network architecture efficiency over generations.

An architecture of a deep neural network can be encoded via two different sets of random variables representing the existence of neurons and synapses in the network. The realization of random variables are binary values, $\{0,1\}$, which determines whether the interested neuron or synapse is realized in the network architecture or not.  However, it is possible to infer the existence of a neuron given the existence of any ingoing or outgoing synapse.

Therefore, the network architecture of a deep neural network can be characterized by $\mathcal{S}$, where \mbox{$\mathcal{S} = \Big\{s^{l,i}\Big\}_{l = 1:L}^{i = 1:I_l}$} is the set of binary states defining the existence of all possible synapses in the network having $L$ possible layers, and $I_l$ possible synapses at each layer~$l$.

The main purpose of evolutionary synthesis frameworks~\cite{EvoNet1} is to model the optimal probability distribution of a network architecture through time and in a generational manner. The generational approach helps the probability distribution of network architecture to account for any change in the network and model the optimal network architecture in a better way. As such, the process of synthesizing a deep neural network is formulated within an evolutionary framework where at each generation a better probability distribution is introduced based on the new network architecture and for the next generation, $g+1$, an offspring deep neural network is synthesized stochastically by a synthesis probability $P(S_g)$:
\begin{align}
\label{eq:mainEvoNet}
P(S_g) = P(\mathcal{S}_g|\mathcal{W}_{g-1}) \cdot F,
\end{align}
\noindent with $P(\mathcal{S}_g|\mathcal{W}_{g-1})$ the synaptic probability model, and $F$ the imposed environmental factors. The offspring networks are then trained at each generation to achieve modeling accuracy while preserving the efficiency and architectural diversity. The environmental factor, $F$, plays the role of prior constraints regarding the desired network architecture and is applied during the synthesis of an offspring network. By use of this approach, the search domain to model the optimal architecture is reduced and as a result, a better probability distribution is created to model the optimal network architecture.  Furthermore, the synaptic probability model $P(S_g)$, which can be treated as the genetic encoding of the network architecture in the context of the evolutionary synthesis framework, plays the main role to evolve a network architecture to survive in the simulated environment $F$.

The effect of genetic encoding on synthesizing efficient network architectures is crucial. Shafiee {\it et al.}~\cite{EvoNet1}  utilized the trained weights of the ancestor network to formulate the genetic encoding and probability distribution.
 Following~\cite{EvoNet1}, the genetic encoding of offspring networks is modeled by $P(\mathcal{S}_g|\mathcal{W}_{g-1})$ where $\mathcal{W}_{g-1}$ represents the set of trained synaptic strengths of the network at generation $g-1$ based on the notion, that the desired traits to be inherited by the offspring networks are related to strong synaptic strengths in the ancestor networks. The synaptic strength of $s_{g-1}^{l,i}$ is represented by $w_{g-1}^{l,i} \in \mathcal{W}_{g-1}$, and the non-existence of a synapse is encoded as $w_{g-1}^{l,i} = 0$ and equals $s_{g-1}^{l,i} = 0$.

\begin{figure}
	\begin{center}
		\includegraphics[width = 8 cm]{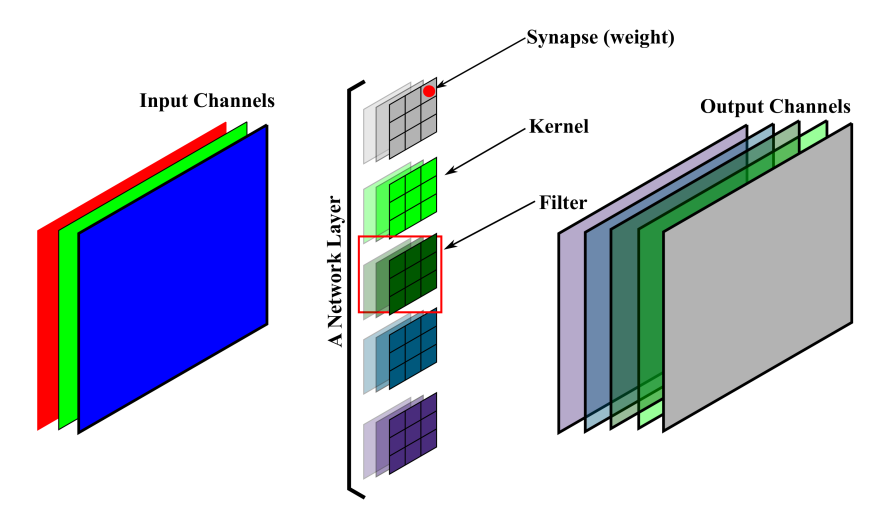}
	\end{center}
	\caption{The visualization of synapse, kernel and filter in a layer of a convolution layer. While a kernel is a 2D structure consisting of synapses, a filter is a combination of kernels with the same size as input channels.  The number of output channels is specified by the number of filters in the layer.  }
	\label{fig:kf}
\end{figure}

A deep neural network is composed of a set of different components in a hierarchical manner where synapses are the smallest components in this hierarchy. As shown in Figure~\ref{fig:kf}, a set of synapses constructs a 2D kernel, a set of kernels creates a filter in a layer and consecutively, a layer in a network is constructed based on different filters.  Due to this hierarchical structure,  Shafiee {\it et al.}~\cite{shafiee2017evolution,shafiee2016evolutionary} further decomposed $P(\mathcal{S}_g|\mathcal{W}_{g-1})$ into a multi-factor probability distribution to promote the formation of synaptic clusters, resulting in the synthesis of offspring deep neural networks that are tailored to be more efficient for computation on parallel computing systems:
\begin{align}
\small
P(\mathcal{S}_g|\mathcal{W}_{g-1}) = \prod_{c \in C}\Big[P(S_{g}^c|\mathcal{W}_{g-1}) \cdot \prod_{(i,l) \in c} P(s_g^{l,i}|w_{g-1}^{l,i})   \Big]
\label{eq:cluster_syn}
\end{align}
\noindent where $S_g^c \subset \mathcal{S}_g$ is a cluster of synapses at generation $g$. A cluster~$c$ can be encoded as a subset of synapses of the network, with a filter or a kernel inside a filter as the examples of synaptic clusters in the genetic encoding scheme~\eqref{eq:cluster_syn}.

The architecture of a deep neural network regarding to the number of synapses, kernels or the numbers of filters  in the network can have different outcomes depends on the host (i.e., computational power) where the execution is taken place. For example, the main speed up for parallel computing devices such as GPUs is when the synthesized network is with less number of filter since a parallel computing device computes a whole filter at the same time. However, this is different for processing on CPU only devices since all processing tasks are computed sequentially.

Therefore, the creation of the genetic encoding $P(\mathcal{S}_g|\mathcal{W}_{g-1})$ is very important as it has a significant influence over the network architectures of offspring deep neural networks in addition to  preserving the accuracy. As mentioned before, the genetic encoding scheme previously proposed is highly dependent on the synaptic strengths of the ancestor deep neural network, i.e., $\mathcal{W}_{g-1}$; therefore, optimizing the distribution of synaptic strengths in $\mathcal{W}_{g-1}$ in a way that promotes optimal genetic encoding $P(\mathcal{S}_g|\mathcal{W}_{g-1})$ favoring the synthesis of offspring neural networks with greater architectural efficiency is highly desired.

Inspired to further improve the architectural efficiency of synthesized offspring deep neural networks, we propose a stress-induced evolutionary synthesis approach where stress signals are imposed on deep neural networks during the training. The stress is encoded as a prior model within a probabilistic framework to induce environmental stresses and better promote the synthesis of robust networks that can achieve greater efficiency while maintaining modeling performance.

\subsection{Stress-Induced Evolutionary Synthesis}
 The general idea is the imposition of epoch-level traumatic stresses to weaken the strengths of a subset of synapses to induce environmental stresses on the network during training. Here, the stresses imposed on the exposed parent deep neural network during training influence the distribution of synaptic strengths in a deep neural network to favor offspring networks with greater architectural efficiency. This effect is transmitted genetically to the next generation, i.e., by  use of the probabilistic genetic encoding. More specifically, the induced stresses encourage configurations of $\mathcal{W}_{g-1}$ that enable more effective genetic encodings $P(\mathcal{S}_{g}|\mathcal{W}_{g-1})$ linked to synthesized offspring networks with greater architectural efficiencies.

Let us model a neural network as a probabilistic model~\cite{weight_uncertainty_2015} $P(y|x;\mathcal{W})$ where $x \in \mathbb{R}^d$ is the $d$-dimensional input to the network, and the network assigns a probability to each possible output $y \in \mathcal{Y}$ regarding the set of trained synaptic strengths $\mathcal{W}$. The learning process of synaptic strengths $\mathcal{W}$ within a deep neural network can be formulated as a maximum likelihood estimation (MLE) given a set of training data $\mathcal{D} = (x_i,y_i)$:
\begin{align}
\hat{\mathcal{W}} &= \underset{\mathcal{W}}{arg\;max}\;\; \log P(\mathcal{D}|\mathcal{W}) \nonumber\\
&= \underset{\mathcal{W}}{arg\;max}\;\; \sum_i \log P(y_i|x_i;\mathcal{W}).
\end{align}
\noindent This optimization is usually performed by a gradient descent approach with the assumption that $\log P(\mathcal{D}|\mathcal{W})$ is differentiable in $\mathcal{W}$.

We now wish to impose prior knowledge to the synaptic strengths $\mathcal{W}$, and re-formulate the problem as a maximum a posteriori (MAP) problem:
\begin{align}
\hat{\mathcal{W}}&= \underset{\mathcal{W}}{arg\;max}\;\; \log P(\mathcal{W}|D) \nonumber \\
& = \underset{\mathcal{W}}{arg\;max}\;\; \log P(D|\mathcal{W}) + \log P(\mathcal{W})
\label{eq:training_MAP}
\end{align}
\noindent where $P(\mathcal{W})$ is the prior model imposed during the training stage. Here, the prior model encodes the imposed stresses to synapses during the training at each generation.  As shown in Figure~\ref{fig:flowdiagram}, we encode the imposed stresses to synapses within the prior model during the training of the deep neural network at each generation. This approach prepares the network for a traumatic change which  would happen on the offspring network and helps ancestor network to transmit such stressful experiences via the probabilistic DNA.

Given the goal that $P(\mathcal{S}_g|\mathcal{W}_{g-1})$ better promotes the synthesis of offspring networks with more effective and efficient network architectures:
\begin{align}
P(\mathcal{S}_g|\mathcal{W}_{g-1}) \approx P \Big(\mathcal{S}_g|\hat{\mathcal{W}}_{g-1}\Big),
\end{align}
\noindent we take advantage of~\eqref{eq:training_MAP} and benefit from $P(\mathcal{W})~:=~P(\hat{\mathcal{W}}_{g-1})$ to provide a more effective genetic encoding scheme.

Here, the prior model, $P({\mathcal{W}}_{g-1})$, is realized as a Binomial probability distribution such that the strengths of a subset of synapses are weakened at each epoch level during the training, and is formulated as follows:
\begin{align}
\hat{\mathcal{W}}_{t+1,g-1} & =\Big [Q_{t,g-1} \ge \bar{U} \Big]\cdot \hat{\mathcal{W}}_{t,g-1} \nonumber\\
& + \beta \cdot  \Big[Q_{t,g-1} < \bar{U} \Big] \cdot \hat{\mathcal{W}}_{t,g-1}
\label{eq:Bernoulli}
\end{align}
\noindent where $Q_{t,g-1}$ is the Binomial distribution formulating such that $P(\hat{\mathcal{W}}_{t,g-1})$, $\bar{U}$ is a set of uniformly distributed random numbers based on uniform distribution $U(0,1)$, $\Big [\cdot\Big]$ is the Iverson bracket determining whether a synapse is selected in~\eqref{eq:Bernoulli} at epoch $t$ for generation $g-1$, and $\hat{\mathcal{W}}_{t,g-1}$ encodes the set of trained synaptic strengths of epoch $t$ at generation $g$. The Binomial distribution $Q_{t,g-1}$ is formulated based on the trained synaptic strengths $\hat{\mathcal{W}}_{t,g-1}$ at epoch $t$:
\begin{align}
Q_{t,g-1} &= q^1_{t,g-1} \cdot q^2_{t,g-1}  \cdot \ldots \cdot q^n_{t,g-1}\\
q^i_{t,g-1} &= \exp(\frac{\hat{w}^i_{t,g-1}}{z^i} -1) \;\;\;\;  1 \leq i \leq n
\end{align}
\noindent where $q^i_{t,g-1}$ is a Bernoulli distribution for the $i$th synapse in a network containing $n$ synapses computed based on $\hat{w}^i_{t,g-1}~\in~\hat{\mathcal{W}}_{t,g-1}$, and $z^i$ a normalization factor.

The factor $0~<~\beta~\leq~1$ is the intra-generational environmental factor applied at each epoch of training to weaken the strength of stochastically selected synapses. The factor $\beta$ imposes minor stress to the deep neural network at the epoch level. These stochastically selected synapses at each epoch are meant to be less important to the modeling power of the deep neural network than other synapses, and weakening them has a minimal effect on the modeling accuracy. However, the cumulation of tiny stress-induced changes shapes the distribution of synaptic strengths to promote the formation of a synaptic probability model $P~\Big(\mathcal{S}_g|\hat{\mathcal{W}}_{g-1}\Big)$ favoring the synthesis of offspring deep neural networks with more efficient yet effective network architectures.  As such, the stress induced by the ancestor deep neural network results in genetic encodings for synthesizing new StressedNets with more efficient yet robust network architectures.

\section{Experimental Results}
\label{sec:exp}

The performance of the proposed stress-induced evolutionary synthesis approach for synthesizing deep neural networks with even greater efficiency while retaining modeling accuracy is evaluated in a comprehensive manner over a variety of different deep neural network architectures for different tasks.  The performance of the proposed approach is also compared with six state-of-the-art methods for achieving efficient deep neural networks~\cite{han2015deep,guo2016dynamic,ullrich2017soft,molchanov2017variational,louizos2017bayesian} on a widely-used benchmark experiment in research literature to provide context in the field.  Furthermore, the effect of the various environmental factors on the quality of the synthesized StressedNets is investigated via parametric analysis.  Finally, a comprehensive qualitative and quantitative feature analysis is performed on the extracted features within the synthesized StressedNets.

\subsection{Efficacy Across Network Architectures}
In this section, the efficacy and generalizability of the proposed stress-induced evolutionary synthesis framework is examined across different network architectures, tasks, and benchmark datasets.

\subsubsection{LeNet5 }
\label{sec:Letnet}
\begin{figure*}
	
	\begin{center}	
		\includegraphics[width = 10 cm]{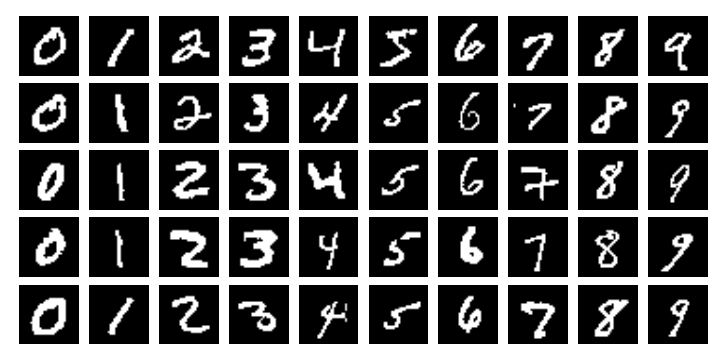}
		\caption{  Example images from the MNIST character recognition dataset.  The MNIST dataset is the combination of 10 digits in $28\times28$ gray scale images. }
		\label{fig:mnist}
	\end{center}
\end{figure*}

For the first experiment, the proposed approach is examined to synthesize StressedNets based on the LetNet5 network architecture for the task of image classification. The original LeNet5 is trained on MNIST dataset~\cite{MNIST}. The MNIST image dataset~\cite{MNIST} comprises of 60,000 training images and 10,000 test images of $28\times28$ handwritten digits (0 to 9). Figure~\ref{fig:mnist} demonstrates some examples from MNIST dataset.  Stress-induced evolutionary synthesis was performed over generations while the error of the synthesized StressedNets is within 1\% of that achieved by the original network.

The LeNet5\footnote{https://github.com/BVLC/caffe/tree/master/examples/mnist} architecture used in this study was implemented in the Caffe platform and is commonly referred to as LeNet5-Caffe. This architecture is comprised of 2 convolutional layers of 20 and 50 filters with $5\times5$ in size and 2 fully connected layers of 800 and 500 neurons.  This variant of LeNet5 was chosen as it is commonly used in research literature for comparing between different methods for achieving efficient deep neural networks.

Table~\ref{Tab:mnist} shows the efficient network architectures and the corresponding classification errors achieved using six state-of-the-art methods~\cite{han2015deep,guo2016dynamic,ullrich2017soft,molchanov2017variational,louizos2017bayesian} and that of two different StressedNets synthesized via the proposed stress-induced evolutionary synthesis approach after 14 and 23 generations, respectively.  It can be observed that the network architecture of the synthesized $StressedNet\#1$ has $\sim 122\times$ fewer weights when compared to the original network architecture while achieving a modeling accuracy of $\sim1.22\%$. Also, it can be observed that the network architecture of the smaller synthesized $StressedNet\#2$ has $\sim270\times$  fewer weights than original network architecture while still achieving a modeling accuracy of $\sim1.64\%$.  More interesting is the fact that $StressedNet\#1$ consists of just \textbf{four} filters in the first convolutional layer and \textbf{seven} filters in the second convolutional layer, while $StressedNet\#2$ contains just \textbf{four} filters in the first convolutional layer and \textbf{five} filters in the second convolutional layer.  This is very important as much of the computational complexity lies in the first two convolutional layers while much of the parameters lie in the fully-connected layers, and as such this significant decrease in the number of filters in convolutional layers within the synthesized StressedNets result in a significant improvement in computational speed.  Taken in the context of performance compared to the other state-of-the-art methods, it can be observed that while the StressedNets have slightly higher error compared to the other methods, their network architectures have comparable number of non-zero weights but have significantly fewer filters in the convolutional layers.  

From the implementation point of view, on parallel processing units, such as GPUs and hardware accelerators on embedded devices, heavy computation of the convolutional layers are the main bottleneck in computational efficiency od deep neural networks compared to the fully connected layers. Therefore, reducing the number of filters in the convolutional layers can significantly speed up the computational and processing time.

\begin{table}
	\begin{center}
		\caption{Experimental results for LeNet5 experiment with MNIST dataset. The network architectures of the synthesized StressedNets are compared with state-of-the-art algorithms for achieving efficient deep neural networks.  The original architecture of LeNet5 is \textbf{20}-\textbf{50}-800-500, where the first two numbers (in bold) shows the number of filters in the convolutional layers while the third and fourth number shows the number of neurons in the fully connected layers. As seen, the proposed framework synthesized two StressedNets (produced after 14 and 23 generations, respectively) that have significantly fewer filters in convolutional layers which is very important for computational efficiency, particularly on parallel computing devices.
		The comparison results were extracted from~\cite{louizos2017bayesian}}
		\label{Tab:mnist}
		\begin{tabular}{lccc}
			Method & $\frac{w \neq 0}{|\mathcal{W}|}$\% & Error\%&Architecture \\ \hline \hline \\
			DC~\cite{han2015deep} & 8.0 & 0.7 & NA\\
			DNS~\cite{guo2016dynamic}&  0.9 & 0.9 & NA \\
			SWS~\cite{ullrich2017soft}&0.5 &1.0 &NA\\ \hline\\
			Sparse VD~\cite{molchanov2017variational}& 0.7& 1.0 &  \textbf{14}-\textbf{19}-242-131 \\
			BC-GNJ~\cite{louizos2017bayesian}&0.9&  1.0 & \textbf{8}-\textbf{13}-88-13\\
			BC-GHS~\cite{louizos2017bayesian} &0.6 &1.0 & \textbf{5}-\textbf{10}-76-16\\
			StressedNet\#1 &0.8 &1.2& \textbf{4}-\textbf{7}-112-24\\
			StressedNet\#2  &0.3 &1.6& \textbf{4}-\textbf{5}-80-12\\
		\end{tabular}
	\end{center}
\end{table}

\subsubsection{AlexNet}

\begin{figure*}
	\begin{center}
			\includegraphics[width = 10 cm]{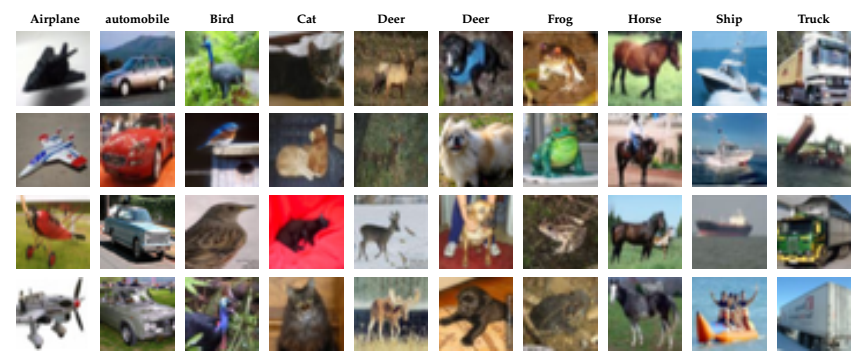}
		
		\caption{ Example images from the CIFAR-10 image classification dataset. CIFAR-10 contains 10 different classes with the images being $32\times32$ in size.  }
		\label{fig:CIFAR}
	\end{center}
\end{figure*}

To examine the efficacy of the proposed stress-induced evolutionary synthesis approach in a larger, more complex deep neural network than LeNet5 for image classification, StressNets were synthesized from the AlexNet network architecture~\cite{krizhevsky2012imagenet}, which consists of 5 convolutional layers of 96, 256, 384, 384 and 256 filters, respectively, and two fully connected layers with 4096 and 4096 neurons. The CIFAR-10~\cite{krizhevsky2009learning} benchmark dataset was used in this experiment, and comprises of 50,000 training and 10,000 test 32$\times$32 natural images (Figure~\ref{fig:CIFAR}) with 10 different classes which equally distributed. To account for the image size in the CIFAR-10 dataset, the kernel sizes of the first two convolutional layers are $5\times5$ and the rest are $3\times3$.  Stress-induced evolutionary synthesis was performed over generations while the error of the synthesized StressedNets is within 2\% of that achieved by the original network.

Table~\ref{Tab:CIFAR} shows a comparison between the network architecture of the original network and the architecture of the synthesized StressedNet (synthesized after eight generations). The network architecture of the synthesized StressedNet is $\sim 40\times$ smaller than that of the original network (i.e., with just 2.5\% of the number of weights compared to the network architecture of the original network), while experiencing a classification error increase of $\sim1.9\%$. It is also important to note that the number of filters in each of the convolutional layers of the StressedNet is more than $7\times$ fewer when compared to that of the original network, and as such the results in noticeably reduced computational complexity particularly on parallel computing devices.  Finally, it can be observed that the model size of the StressedNet is $\sim37\times$ smaller when compared to that of the original network, which is very beneficial for embedded scenarios where the memory and storage are limited.

\begin{table*}
	\begin{center}
		\caption{Experimental results for the AlexNet experiment with the CIFAR-10 dataset. The network architecture of the synthesized StressedNet (synthesized after 8 generations) is compared with that of the original network.  The synthesized StressedNet consists of only 2.5\% of the number of non-zero weights compared to that of the original network and has a model size  $\sim38\times$ smaller model size than  the original network.  The bold-face numbers show the number of filters in convolutional layers. It can be observed that the StressedNet's network architecture has $6\times$ fewer filters when compared to the original AlexNet network, which leads to significant improvement in computational efficiency.    }
		\label{Tab:CIFAR}
		\begin{tabular}{l||c|c|c|c}
			Method & Model Size (MB) &$\frac{w \neq 0}{|w|}$\% & Error\%&Architecture \\  \hline \hline
			AlexNet &30.979 & 100&13.4 &\textbf{96}-\textbf{256}-\textbf{384}-\textbf{384}-\textbf{256}-4096-4096 \\
			StressedNet&  0.820& 2.5&15.3 &\textbf{18}-\textbf{39}-\textbf{58}-\textbf{60}-\textbf{40}-640-64 \\
		\end{tabular}
	\end{center}	
\end{table*}

\subsubsection{YOLOv2}
One of the main advantages of the proposed stress-induced evolutionary synthesis framework is that it is easily generalizable to a variety of different network architectures for different tasks.  To demonstrate this flexibility, StressedNets were synthesized using the proposed approach based on the YOLOv2 network architecture~\cite{redmon2016yolo9000} for the task of object detection.  In YOLOv2, the object detection problem is formulated as a single regression problem and bounding box coordinates and class probabilities are computed at the same time.  As a result, besides achieving state-of-the-art performance, the network architecture of YOLOv2 is considerably smaller and more efficient when compared to other object detection deep neural networks and thus is considered as one of the fastest object detection approaches in research literature. Despite its efficient design, it is still currently not tractable to achieve reasonable inference speed using YOLOv2 on embedded systems for a number of operational scenarios.

\begin{figure*}

	\begin{center}
		\setlength{\tabcolsep}{0.8mm}
			\includegraphics[width = 14 cm]{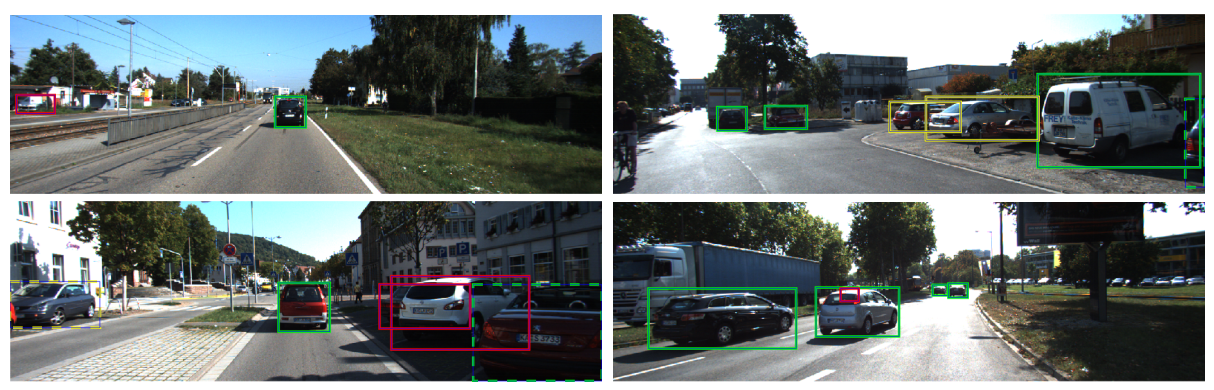}	
		\caption{Example images from the KITTI object detection dataset.  Ground truth object bounding boxes are overlaid onto the images for context.}
		\label{fig:Kitti}
	\end{center}
\end{figure*}

To investigate the efficacy of the proposed framework for synthesizing StressedNets to perform fast object detection on embedded systems, a YOLOv2 network was trained on the KITTI benchmark dataset for the purpose of detecting two types of objects in an image: i) car, and ii) pedestrian.  Figure~\ref{fig:Kitti} shows some examples from the KITTI dataset. For each of the two object types (car and pedestrian) used in this experiment from the KITTI benchmark dataset, there are three different task difficulty groups: i) Easy\footnote{\textbf{Easy}: Min. bounding box height: 40 Px, Max. occlusion level: Fully visible, Max. truncation: 15\%}, ii) Moderate\footnote{\textbf{Moderate}: Min. bounding box height: 25 Px, Max. occlusion level: Partly occluded, Max. truncation: 30\%}, and iii) Hard\footnote{\textbf{Hard}: Min. bounding box height: 25 Px, Max. occlusion level: Difficult to see, Max. truncation: 50\%}.  In this experiment, stress-induced evolutionary synthesis was performed over generations while the average precision (AP) of the synthesized StressedNets is within 2\% of that achieved by the original network.

The proposed framework synthesized an evolved network architecture after 70 generations which is $33.43\times$ smaller in size than original YOLOv2 architecture (i.e., 5.5 MB compared to 184 MB).

Table~\ref{Tab:YOLO} shows a comparison between the network architecture of the original network and the architecture of the synthesized StressedNet (synthesized after 70 generations).  It can be clearly observed that the synthesized StressedNet provides comparable results and in some cases outperforms the original YOLOv2 network in terms of AP for both object types under all difficulty scenarios, and yet is $\sim33.45\times$ smaller in model size when compared to that of the original network (i.e., 5.5 MB compared to 184 MB).  This experiment illustrates the over-parameterization issues faced in deep neural networks when faced with limited training data sizes or a task that does not require the full information capacity of the network. As seen, the proposed framework synthesizes highly efficient deep neural networks that possess network architectures with sufficient information capacity for the task at hand.

For the last experiment in this section, the inference speed of both the original network and the synthesized StressedNet were examined on the Nvidia Tegra X1 mobile processor.  As seen in Table~\ref{Tab:YOLO}, it can be observed that while the original YOLOv2 network achieved 2.45 frames per second, the synthesized StressedNet achieved 13.43 frames per second (a $5.5\times$ speed-up comapred to the original network).

\begin{table*}
	\begin{center}
		\caption{Experimental results for YOLOv2 experiment with KITTI dataset. The network architecture of the synthesized StressedNet (synthesized after 70 generations) is compared with that of the original network.  The synthesized StressedNet had a model size that was $\sim33.45\times$ smaller than that of the original network yet outperformed the original network in all test scenarios.  It can be observed that the inference speed of the synthesized StresssedNet on the Nvidia Tegra X1 mobile processor was $\sim13.43$ FPS, which is $\sim5.5\times$ faster than that of the original YOLOv2 network.}
		\label{Tab:YOLO}
		\begin{tabular}{l||c||c||ccc||ccc}
			Method & Model Size (MB) & Inference speed (FPS) &\multicolumn{3}{c}{AP Car (\%)} & \multicolumn{3}{c}{AP Pedestrian (\%)} \\
			~&~&~ & \textbf{Easy}&\textbf{Moderate}&\textbf{Hard}& \textbf{Easy}&\textbf{Moderate}&\textbf{Hard}\\ \hline \hline
			YOLOv2 & 184 & 2.45 &93.47 & 85.19&77.54&78.78 &76.84&69.46  \\
			StressedNet& 5.5  & 13.43 &94.72&86.06&79.16&78.87&75.55&68.96 \\
		\end{tabular}
	\end{center}	
\end{table*}

\subsection{Effect of Environmental Factors}

The effect of different environmental factors and different levels of enforcement of $F$ in \eqref{eq:mainEvoNet} on synthesizing  more efficient network architectures  is studied in this section. To analyze  the behavior of the synthesized StressedNets through several generations, the YOLOv2 network architecture is utilized as the original ancestor network architecture for three different experiments with three different environmental factor values using the KITTI dataset.

New offspring StressedNets are synthesized through 5 consecutive generations, and based on six different environmental factors $\{0.95, 0.9, 0.8, 0.6, 0.4, 0.2\}$ (which indicate the expected ratio between the number of synapses in a descendant offspring network and that of its direct ancestor network as enforced during the stressed-induced evolutionary synthesis process).

\begin{figure*}
	\begin{center}

		\includegraphics[scale =0.6]{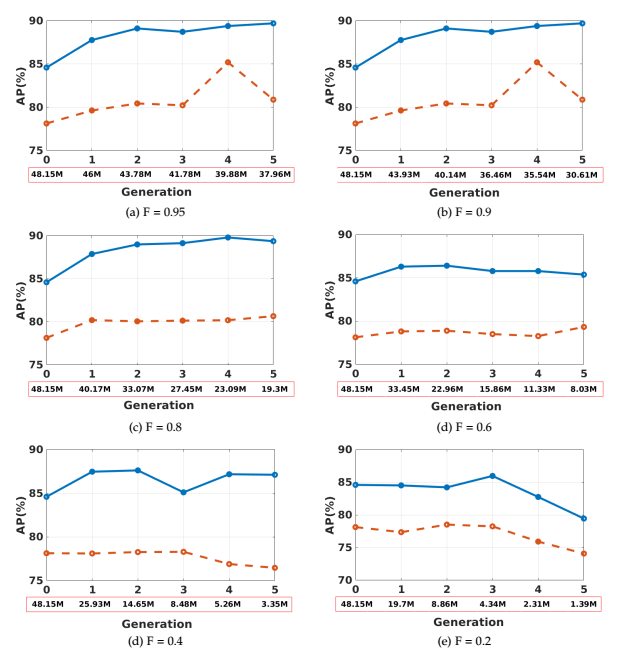}
		\caption{ Environmental factor analysis. The effect of different environmental factors (i.e., $F$ in \eqref{eq:mainEvoNet}) is studied by performing stressed-induced evolutionary synthesis over generations using the YOLOv2 network architecture for six different factor values. ``Generation 0" demonstrates the original network architecture statistics. }
		\label{fig:genEval}
	\end{center}
\end{figure*}

Figure~\ref{fig:genEval} illustrates the modeling accuracy of StressedNets (in this case, AP for cars and pedestrians for the Moderate difficulty scenarios) at different generations for each of the six different environmental factors.   It can be observed that the synthesized StressedNets produced via all three tested configurations (i.e., environmental factors 0.95, 0.9, 0.8) provide better accuracy compared to the original YOLOv2 network. However when the environmental factor is decreased (i.e., 0.6, 0.4 and 0.2) the resulted StressedNet models provide comparable and sometimes worse modeling accuracy compared to the original YOLOv2 network.  This result demonstrates that when network facing severe conditions it cannot survive and maintain its modeling accuracy.   

Figure~\ref{fig:genEval}(a) shows the performance of synthesized StressedNets based on an environmental factor of 0.95. After five generations, the network architecture of the synthesized StressedNet had $\sim$9 million fewer parameters while outperforming the original network by $\sim 5 \%$.

Figure~\ref{fig:genEval}(b) shows the performance of synthesized StressedNets based on an environmental factor of 0.90. It can be observed that by enforcing greater efficiency in descendent offspring networks, the StressedNets synthesized here had noticeably higher efficiency at the same generation when compared to the last experiment (for example, after five generations the network architecture had $\sim$18 million fewer parameters), while still outperforming the original YOLOv2 by $\sim 5 \%$.

Figure~\ref{fig:genEval}(c) shows the performance of synthesized StressedNets based on an environmental factor of 0.80. It can be observed that even at this aggressive level of efficiency enforcement the trend persists where the StressedNets synthesized here had noticeably higher efficiency at the same generation when compared to the last experiment (for example, after five generations the network architecture had half the number of parameters).

Figure~\ref{fig:genEval}(d), (e) and (f) demonstrate the performance of the synthesized StressedNet architectures when being imposed with more severe environmental factors (i.e., environmental factors, 0.6, 0.4 and 0.2). As seen when the offspring networks are enforced in such severe environments, the networks cannot maintain their modeling accuracies and it results to lose their modeling accuracy over generations. This effect can be illustrated by comparing the generation 6 in the experiment 3 (i.e., F=0.8 Figure~\ref{fig:genEval}(c)) and generation 2 in the experiment 6 (i.e., F=0.2 Figure~\ref{fig:genEval}(f)). As seen, the offspring StressedNet in the experiment 3 provides $\sim$5\% increase in modeling accuracy compared to the original YOLOv2 while the synthesized StressedNet in the experiment 6 has slight drop in the modeling accuracy compared to the original YOLOv2 network even with the larger numbers of parameters compared to generation 6 in the experiment 3 (i.e.,F=0.8 Figure~\ref{fig:genEval}(c)).
      
The conducted experiments illustrate that while it is possible to synthesize deep neural networks with highly efficient network architectures with the proposed stress-induced evolutionary synthesis approach, while achieving modeling accuracies comparable and in some cases higher than the original network. However, it is important to consider that  while choosing smaller environmental factors produces more efficient networks in less number of generations, the offspring networks cannot retain the complete modeling accuracy in this situations. Therefore, there is a trade-of between the processing time for synthesizing  efficient networks and synthesizing the best possible efficient network architecture.          

\subsection{Feature Analysis}

As the last experiment of this study, a comprehensive feature analysis on the generated feature representations within StressedNets is performed both qualitatively (i.e., visually) and quantitatively to investigate their discriminative capacity.

\begin{figure*}
 	\begin{center}
 			\includegraphics[scale =0.8]{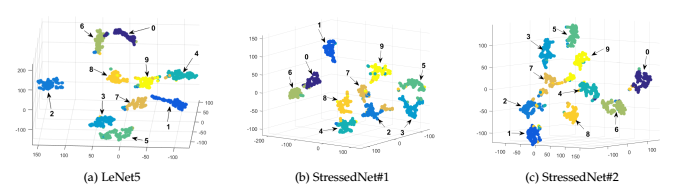}
 		\caption{ t-SNE visualization for extracted features from original LetNet5 network compared to those from two different StressedNets. As seen, while the between-class distances decrease in the feature domains formed by the StressedNets, they still maintain the separability among the classes.     }
 		\label{fig:tsne}
 	\end{center}
 \end{figure*}

To qualitatively investigate the discriminative capacity of the generated feature representations within StressedNets, a deep neural network based on the LeNet5 architecture mentioned in Section~\ref{sec:Letnet} was trained on the MNIST dataset and stressed-induced evolutionary synthesis was performed on that network.  The extracted features from the last fully connected layer of two StressedNets at different generations, along with that of the original network, were extracted and t-SNE~\cite{maaten2008visualizing} was performed on the features to visualize the extracted features (see Figure~\ref{fig:tsne}).  t-SNE is a variation of stochastic neighbor embedding which visualizes high-dimensional data by giving each datapoint a location in a 2- or 3D map.  The algorithm is capable of capturing much of the local structure of the high-dimensional data very well and it is optimized much easier compared to other state-of-the-art visualization algorithms. The ability to preserve the local structure is an important aspect as we want to study high-dimensional feature representations produced by deep neural networks.  To have a better visualization,  100 samples for each class in the MNIST dataset were selected for visualization purposes.

Each network produces a different number of features based on the size of the last fully connected layer; as seen in Table~\ref{Tab:mnist} the original LeNet5 network produces 500 features while the two StressedNets (i.e., StressedNet\#1 and StressedNet\#2) results in 26 and 12 features, respectively.

As seen in Figure~\ref{fig:tsne}, while the between-class distances of the feature representations in some specific classes decrease in the feature domains formed by the synthesized StressedNets, they still maintain the separability among the classes. Figure~\ref{fig:tsne} shows that classes $\{0, 1, 2, 3, 4 \}$ were affected the least among all classes and classes $\{6, 8, 9\}$ were dispersed the most. To have a better understanding of the separability between classes, \mbox{1-NN} classification was conducted for all test samples based on 100 selected training samples (i.e., a same set of training for all experiments). Table~\ref{Tab:Class} shows the \mbox{1-NN} classification accuracy for all 10 classes in the MNIST dataset. The results show a consistent trend on classification accuracies where the classification accuracies for classes $\{0, 1, 2, 3, 4 \}$ are almost the same in the original network and the synthesized StressedNets while the larger difference in classification accuracies took place in classes  $\{6, 8, 9\}$.

Table~\ref{Tab:Class} also demonstrates the results of \mbox{1-KNN} when PCA~\cite{wold1987principal} was applied on the generated features within the original LeNet5 network. To show the effectiveness of the StressedNets and demonstrate that their new network architectures not only shrink the feature vectors but also creates a new highly discriminative feature space for the input data, two new feature sets with the same size as the generated features within StressedNet\#1 (i.e., 26 features) and StressedNet\#2 (i.e., 12 features) were extracted via PCA as well. The same \mbox{1-NN} performance metric was conducted to compare the effectiveness of generated features within the StressedNets. As seen in Table~\ref{Tab:Class}, I) the comparison of modeling accuracies shows that  the discriminatory power of the StressedNet feature representations are higher than that obtained using just PCA applied on the features extracted by the original LeNet5. II) Comparing  the total classification accuracies of the original LetNet5 and StressedNet illustrates  the discriminatory power of the StressedNet feature representations which are comparable to that of LeNet5 and shows the information density of the StressedNet features has the same level as the original LeNet5 architecture.

\begin{table*}
	\begin{center}
		\caption{1-NN modeling accuracy; The extracted features by the original LeNet5 network and the two StressedNets are compared by the 1-NN classification procedure.} 
		\label{Tab:Class}
		\begin{tabular}{l||cccccccccc|c}
			~&\multicolumn{10}{c}{Class Label(\%)}&~\\
			Model & ``0"& ``1"&``2"&``3"&``4"&``5"&``6"&``7"&``8"&``9" & Total\\  \hline \hline
			LetNet5 &0.9928&0.9982&	0.9835&	0.9801&	0.9887&	0.9820&	0.9906&	0.9834&	0.9856&	0.9861&0.9873\\
			StressedNet\#1& 0.9938&	0.9938&	0.9825&	0.9762&	0.9867&	0.9719&	0.9864&	0.9785&	0.9774&	0.9772&0.9827\\	
			StressedNet\#2 & 0.9887&	0.9920&	0.9777&	0.9782&	0.9745&	0.9641&	0.9749&	0.9737&	0.9640&	0.9702&0.9762  \\  \hline \hline
			PCA-26-Features & 0.9755&	0.9938&	0.9437&	0.9524&	0.9714&	0.9562&	0.9728&	0.9328&	0.9661&	0.9554&	0.9623\\
			PCA-16-Features &0.9785&	0.9903&	0.9515&	0.9485&	0.9786&	0.9506&	0.9613&	0.9319&	0.9496&	0.9544& 0.9599

		\end{tabular}
	\end{center}	
\end{table*}

\section{Conclusion}
In this paper, a stress-induced evolutionary synthesis framework is proposed for synthesizing progressively more efficient deep neural networks over generations. By inducing stresses upon networks during the training stage within the evolutionary synthesis framework, the robustness of the synthesized networks in facing traumatic changes is improved, which as a consequence promotes the synthesis of descendant deep neural networks with improved model fidelity at a greater efficiency. Experimental results across a variety of different network architectures and datasets demonstrate the effectiveness and generality of the proposed stress-induced evolutionary synthesis framework in synthesizing more efficient deep neural networks while preserving the modeling accuracy, which makes them very well-suited for use in embedded applications where the computational and memory resources are highly limited.

\ifCLASSOPTIONcompsoc
m
  \section*{Acknowledgments}

\else
  \section*{Acknowledgment}
\fi

  The authors would like to thank ADC Automotive Distance Control Systems GmbH Continental, the Canada Research Chairs program, and Natural Sciences and Engineering Research Council of Canada (NSERC) for their financial support. The authors also thank Nvidia for the GPU hardware used in this study through the Nvidia Hardware Grant Program.

\ifCLASSOPTIONcaptionsoff
  \newpage
\fi


\end{document}